\documentclass{llncs}

\usepackage[utf8]{inputenc}
\usepackage{graphicx}
\usepackage[english]{babel}

\newcommand{\items}[0]{{\cal I}}
\newcommand{\lang}[1]{{\cal L}_{#1}}
\newcommand{\freq}{freq}
\newcommand{\minfr}[0]{\textit{minfr}}
\newcommand{\maxfr}[0]{\textit{maxfr}}
\newcommand{\transactions}[0]{{\cal T}}

\title{Discovering Knowledge using a Constraint-based Language}

\author{Patrice Boizumault, Bruno Crémilleux, Mehdi Khiari, \\ Samir Loudni, and Jean-Philippe Métivier}
\institute{
University of Caen Basse-Normandie -- GREYC (CNRS UMR 6072)\\
Campus II Cote de Nacre, 14000 Caen - France\\
\texttt{\{firstname.lastname\}@unicaen.fr}
}

\begin{document}

\maketitle

\bigskip
{\noindent {\bf DPA 11201} -- This work was presented into the Dagstuhl
  Seminar "Constraint Programming meets Machine Learning and Data Mining"
  organized by Luc De Raedt, Heikki Mannila, Barry O'Sullivan, and Pascal
  Van Hentenryck - May 15-20, 2011.}

\medskip

\begin{abstract}
  Discovering pattern sets or global patterns is an attractive issue from
  the pattern mining community in order to provide useful information. By
  combining local patterns satisfying a joint meaning, this approach
  produces patterns of higher level and thus more useful for the data analyst
  than the usual local patterns, while reducing the number of patterns. In
  parallel, recent works investigating relationships between data mining
  and constraint programming (CP) show that the CP paradigm is a nice
  framework to model and mine such patterns in a declarative and generic
  way. We present a constraint-based language which enables us to define
  queries addressing patterns sets and global patterns. The usefulness of
  such a declarative approach is highlighted by several examples coming
  from the clustering based on associations. This language has been
  implemented in the CP framework.
\end{abstract}

\medskip

\section{Introduction}

%

Over the two last decades, local pattern discovery has became a rapidly
growing field~\cite{MBS05lpd} and several paradigms are available for
producing extensive collections of patterns such as the constraint-based
pattern mining~\cite{NLHP98}, condensed representations of
patterns~\cite{CRB05}, interestingness measures~\cite{GengH06} as well as
integrating external resources and background
knowledge~\cite{LavracZD05LPintegratingBK}. Because of the exhaustive
nature of the techniques, the pattern collections provide a fairly complete
picture of the information content of the data. However, this approach
suffers from limitations. First, the collections of patterns still remain
too large for an individual and global analysis performed by the data
analyst. Secondly, the so-called local patterns represent fragmented
information and patterns expected by the data analyst require to consider
simultaneously several local patterns. In this work, we propose a
declarative approach addressing the issue of discovering patterns combining
several local patterns.


The data mining literature includes many methods to take into account the
relationships between patterns and produce global patterns or pattern
sets~\cite{DRZ07a,GiacomettiMMSideal09pbm}. Recent approaches -
constraint-based pattern set mining \cite{DRZ07a}, pattern teams
\cite{KH06a} and selecting patterns according to the added value of a new
pattern given the currently selected patterns \cite{BZ07a} - aim at
reducing the redundancy by selecting patterns from the initial large set of
local patterns on the basis of their usefulness in the context of the other
selected patterns.  Even if these approaches explicitly compare patterns,
they are mainly based on the reduction of the redundancy or specific aims
such as classification processes. Heuristic functions are often used and
the lack of methods to mine complete and correct pattern sets or global
patterns may be explained by the difficulty of the task. Mining local
patterns under constraints requires the exploration of a large search space
but mining global patterns under constraints is even harder because we have
to take into account and compare the solutions satisfying each pattern
involved in the constraints.  The lack of generic approaches restrains the
discovery of useful global patterns because the user has to develop a new
method each time he wants to extract a new kind of global patterns. It
explains why this issue deserves our attention.

In this paper, we propose a constraint-based language to discover patterns
combining several local patterns. The data analyst expresses his/her
queries thanks to constraints over terms built from constants, variables,
operators, and function symbols. The key idea is to propose a generic and
declarative approach to ask queries: the user models a problem by
specifying a set of constraints and then a Constraint Programming (CP)
system is responsible for solving it. This work is in the spirit of the
cross-fertilization between data mining and CP which is a research field in
emergence~\cite{DBLP:journals/ai/GunsNR11,Khiari-09,CI-KHIARI-10b,RI-KHIARI-11,DBLP:conf/kdd/RaedtGN08,DRGN10}.

The constraint-based language offers the great advantage to provide an easy
method to address different problems: it is enough to change the
declarative specification in term of constraints. We illustrate the
approach by several examples coming from the clustering based on
associations: with simple query refinements, the data analyst is able to
easily produce clusterings satisfying different properties.  We think that
the process greatly facilitates the building of global patterns and the
discovery of knowledge. We do not detail in this paper the solving step, a
preliminary implementation of the constraint-based language is given
in~\cite{CI-KHIARI-10b}.

\medskip

This paper is organized as follows. Section~\ref{sec:cbl} describes the
constraint-based language and shows how queries and constraints can be
defined using terms and built-in constraints. Starting from the clustering
example, Section~\ref{from_modelling} depicts the process of successive
refinements which enables us to easily address several kinds of clustering and
then the discovery of global models.


\newpage

\section{A Constraint-based Language}

\label{sec:cbl}

In this section, we describe the constraint-based language we propose.
Terms are built using constants, variables, operators, and function symbols.
Constraints are relations over terms that can be satisfied or not.
First, we recall definitions.
Then, we describe terms and present how the data analyst can define new function symbols using operators and built-in function symbols.
Finally, we introduce constraints and show how queries and constraints can be defined using terms and built-in constraints.

\subsection{Definitions and example}

Let $\items$ be a set of $n$ distinct literals called items, an itemset
(or \emph{pattern}) is a non-null subset of $\items$. The language of itemsets
corresponds to $\lang{\items} = 2^{\items} \backslash \emptyset$. A
\emph{transactional dataset} is a multi-set of $m$ itemsets of $\lang{\items}$. Each
itemset, usually called a \emph{transaction} or object, is a database entry. 
For instance, Table~\ref{tab} gives a transactional dataset~$\transactions$ where $m$$=$$11$
transactions $t_1, \dots, t_{11}$ are described by $n$$=$$8$ items $A, B, C, D, E, F, G, H$.

\smallskip
\noindent{\bf Definition 1. (frequency)}
The frequency of a pattern is the number of transactions it covers.
Let $X_i$ be 
a pattern, \texttt{freq}$(X_i)$ $=$ $\mid$$\{t \in \transactions \mid X_i \subseteq t \}$$\mid $.
 
So, \texttt{freq}$(\{A, E\}) = 3$ and \texttt{freq}$(\{C, F, G, H\}) = 1$.
The \emph{frequency} constraint focuses on patterns occurring in the dataset a number of times exceeding a given minimal threshold: \texttt{freq}$(X_i)$ $\ge \minfr$.
An other interesting measure to evaluate the relevance of patterns is the area~\cite{GGTds04}.

\smallskip
\noindent{\bf Definition 2. (area)}
Let $X_i$ be 
a pattern, \texttt{area}$(X_i)$ $=$ \texttt{freq}$(X_i)$ $\times$ \texttt{size}$(X_i)$ where \texttt{size}$(X_i)$ denotes the cardinality of $X_i$.

For transactional dataset $\transactions$ (see Table \ref{tab}), there are nine patterns satisfying the constraint \texttt{area}$(X) \ge 6$~: 
$\{A, E, G\}$, $\{B, E, G\}$, $\{C, E, G\}$, $\{C, E, H\}$, $\{E, G\}$, $\{C, E\}$, $\{C, H\}$, $\{E\}$, $\{G\}$.

\begin{table}[t]
\begin{center} 
{\small
\begin{tabular}{c|cccccccc}
Trans. & \multicolumn{8}{|c}{Items} \\
\hline
$t_1$    & A &   &   & D &   & F &   &  \\
$t_2$    & A &   &   &   & E & F &   &  \\
$t_3$    & A &   &   &   & E &   & G &  \\
$t_4$    & A &   &   &   & E &   & G &  \\
$t_5$    &   & B &   &   & E &   & G &  \\
$t_6$    &   & B &   &   & E &   & G &  \\
$t_7$    &   &   & C &   & E &   & G &  \\
$t_8$    &   &   & C &   & E &   & G &  \\
$t_9$    &   &   & C &   & E &   &   & H \\
$t_{10}$ &   &   & C &   & E &   &   & H \\
$t_{11}$ &   &   & C &   &   & F & G & H \\
\hline
\end{tabular}
}
\vspace{0.15cm}
\caption{Transactional dataset $\mathcal{T}$.} \label{tab}
\end{center}
\end{table}

\subsection{Terms}

Terms are built using:
\begin{enumerate}
\item 
{\bf constants} are either numerical values (as threshold $\minfr$),
or items (as $A$)
or patterns (as $\{A, B\}$)
or transactions (as $t_7$).				
\item 
{\bf variables}, noted $X_i$, for  $1 \leq i \leq k$, represent the unknown patterns.
\item 
{\bf operators}:
\begin{itemize}
\item set operators as $\cap, \cup, \backslash$, \ldots
\item numerical operators as $+, -, \times$, $/$, \ldots
\end{itemize}
\item 
{\bf function symbols} involving one or several patterns: 
\texttt{freq/1}, \texttt{size/1}, \texttt{cover/1}, \texttt{overlapItems/2},  \texttt{overlapTransactions/2}, \ldots
\end{enumerate}

Terms are built using constants, variables, operators, and function symbols.
Examples of terms:
\begin{itemize}
\item  $\texttt{freq}(X_1) \times \texttt{size}(X_1)$
\item  $\texttt{freq}(X_1 \cup X_2) \times \texttt{size}(X_1 \cap X_2)$
\item  $\texttt{freq}(X_1) - \texttt{freq}(X_2)$
\end{itemize}

\subsubsection{i) Built-in function symbols.} 
\label{functeurs}

Our constraint based language owns predefined (built-in) function symbols\footnote{Only function symbols used in Section \ref{from_modelling} 
are introduced in this paper.} like:

\begin{itemize}
\item 
\texttt{cover}$(X_i) = \{t \mid t \in \mathcal{T}, X_i \subseteq t \}$ is the set of transactions covered by $X_i$.
\item 
\texttt{freq}$(X_i) = \, \mid \{t \mid t \in \mathcal{T}, X_i \subseteq t \} \mid$ 	
\item 
\texttt{size}$(X_i) = \, \mid \{j \mid j \in \items, j \in X_i \} \mid$ 				
\item 
\texttt{overlapItems}$(X_i, X_j) = \, \mid X_i \cap X_j \mid$ is the number of items shared by both $X_i$ and $X_j$.
\item 
\texttt{overlapTransactions}$(X_i, X_j) = \, \mid\!\texttt{cover}(X_i) \cap \texttt{cover}(X_j) \mid$ is the number of transactions covered by both $X_i$ and $X_j$.
\end{itemize}

\subsubsection{ii) User-defined function symbols.} 

The data analyst can define new function symbols using constants, variables, operators 
and existing function symbols (built-in or previously defined ones).
Examples:
\begin{itemize}
\item $\texttt{area}(X_i) = \texttt{freq}(X_i) \times \texttt{size}(X_i)$
\item $\texttt{coverage}(X_i, X_j) = \texttt{freq}(X_i \cup X_j) \times \texttt{size}(X_i \cap X_j)$
\item Let $D_1$, $D_2 \subset \transactions$ be 2 sets of transactions and \texttt{freq}$(X_i, D_j)$ the frequency of pattern $X_i$ into $D_j$, then:
$$
\texttt{growth-rate}(X_i) = { \frac{ \mid D_2 \mid \times \, \texttt{freq}(X_i, D_1)}{\mid D_1 \mid \times \, \texttt{freq}(X_i, D_2)} }\\
$$		
\end{itemize}

\subsection{Constraints and Queries}
\label{contraintes}

Constraints are relations over terms.
They can be either {\it built-in} or  {\it user-defined}.
There are three kinds of constraints:

\begin{enumerate}
\item 
\textbf{numerical} ones like: $<$, $\leq$, $=$, $\neq$, $\geq$, $>$, \ldots \\
Examples:
\begin{itemize}
\item $\texttt{freq}(X_1) \leq 10$
\item $\texttt{size}(X_2) = 2 \times \texttt{size}(X_3)$
\item $\texttt{area}(X_1) < \texttt{size}(X_2) \times \texttt{size}(X_3)$
\end{itemize}
\item 
\textbf{set} ones like: $=$, $\neq$, $\in$, $\notin$, $\subset$, $\subseteq$, \ldots \\
Examples:
\begin{itemize}
\item $i_3 \in X_1$
\item $X_1 \cup X_2 \subset X_3$
\item $X_1 = X_2 \cap X_4$
\end{itemize}
\item
\textbf{dedicated} ones like:
\begin{itemize}
\item 
\texttt{closed}$(X_i)$ is satisfied iff $X_i$ is a closed\footnote{Let $Tr_i$ be the set of transactions covered by pattern $X_i$. 
$X_i$ is closed iff $X_i$ is the largest ($\subset$) pattern covering $Tr_i$.} pattern. 
\smallskip
\item 
\texttt{coverTransactions}$([X_1, ..., X_k])$ is satisfied iff each transaction is covered by at least one pattern (i.e. $\bigcup_{1 \le i \le k}  \texttt{cover}(X_i)$ = $\mathcal{T}$),
\smallskip
\item 
\texttt{coverItems}$([X_1, ..., X_k])$ is satisfied iff every item belongs to at least one pattern (i.e. $\bigcup_{1 \le i \le k} X_i$ = $\items$).
\smallskip
\item 
\texttt{canonical}$([X_1, ..., X_k])$ is satisfied iff for all $i$ s.t. $1 \leq i < k$, pattern $X_i$ is less than pattern $X_{i+1}$ with respect to the lexicographic order.
\end{itemize}
\end{enumerate}

Queries and constraints are formulae built using constraints and logical connectors: $\wedge$ (conjunction) and $\vee$ (disjunction).

In the following, we take the exception rules as example\footnote{For more examples, see the modelling  of the clustering problem (Section~\ref{from_modelling}).}.
An \textit{exception rule}\footnote{The definition of exception rules initially presented in \cite{Suzuki02} also includes a reference rule $X_2 \not\rightarrow \neg I$.} is  a pattern combining a strong rule and a deviational pattern to the strong rule:

\begin{displaymath}
e(X_1\rightarrow\neg I) \equiv
\left\{ \begin{array}{l l}
true & \textrm{if } \exists X_2 \in \lang{\items} \textrm{ such that }  X_2
  \subset X_1, \textrm{one have } \\
    & \hspace*{2cm} (X_1 \backslash X_2\rightarrow I) \wedge (X_1 \rightarrow \neg I)\\ 
 false & \textrm{otherwise}
 \end{array} \right.
 \end{displaymath}

%
adding $X_2$ to $X_1 \backslash X_2$ provides the exception rule $X_1 \to \neg I$

\begin{itemize}
\item $X_1\backslash X_2\rightarrow I$ must be a frequent rule having a high
  confidence value:
\item $ X_1\rightarrow\neg I$ must be a rare rule having a high confidence
  value:
\end{itemize}

to sum up:
\begin{displaymath}
exception(X_1,X_2) \equiv\left\{ \begin{array}{ll}
\freq((X_1\setminus X_2)\sqcup I)\geq\minfr\ \wedge\\
(\freq(X_1\setminus X_2)-\freq((X_1\setminus X_2)\sqcup I))\leq\delta_{1}\ \wedge\\
\freq(X_1\sqcup\neg I)\leq \maxfr\ \wedge\\
(\freq(X_1)-\freq(X_1\sqcup\neg I))\leq\delta_{2}
\end{array} \right.
\end{displaymath}

\section{From Modelling to Solving}
\label{from_modelling}

The major strength of our approach is to provide a simple and efficient way to refine a query.
In practice, the data analyst begins with submitting a first query $Q_0$.
Then, he will successively refine this query (deriving $Q_{i+1}$ from $Q_i$)
until he considers that relevant information has been extracted.


Clustering models aim at partitioning data into groups (clusters) 
so that transactions occurring in the same cluster are similar 
but different from those appearing in other clusters.
We selected the clustering problem to illustrate our approach for two main reasons.
First, clustering is an important and popular unsupervised learning method \cite{Berkhin02surveyof,DBLP:journals/ml/Fisher87,DBLP:conf/vldb/GibsonKR98}.
Then, by nature, clustering proceeds by iteratively refining queries until a satisfactory solution is found.
The clustering model, used here, starts from closed patterns because a closed pattern is a pattern gathering the maximum amount of similarity between a set of transactions.

\subsection{Modelling a clustering query}
\label{pbsol}

The usual clustering problem can be defined as follows: 
\begin{center}
``to find a set of $k$ closed patterns $X_1, X_2, ..., X_k$ covering all transactions without any overlap on these transactions''.
\end{center}

First, \texttt{closed}$(X_i$) constraints (see Section \ref{contraintes}) are used to enforce each unknown pattern $X_i$ to be closed, 

Then, it is easy to constrain the set of patterns to cover all the transactional dataset
using the \texttt{coverTransactions}$[X_1, X_2, .., X_k])$ constraint (see Section \ref{contraintes}).

Finally, to avoid any overlap over the transactions, for each couple of patterns $(X_i, X_j), i<j$, 
a constraint \texttt{overlapTransactions}($X_i, X_j$)$=0$ is added.
This constraint states that there is no transaction covered by both $X_i$ and $X_j$.

The following query ($Q_0$) models the initial clustering problem:
$$
\left\{
\begin{array}{l}
\land_{1 \leq i \leq k}~\texttt{closed}(X_i)~\land\\
\texttt{coverTransaction}([X_1,...,X_k])~\land\\
\land_{1 \leq i < j \leq k}~\texttt{overlapTransactions}(X_i, X_j) = 0\\
\end{array}
\right .
$$

On our running example, when looking for a clustering with $k=3$ patterns, we obtain $30$ solutions (See Table \ref{sols30}). 

\begin{table}[t]
\begin{center}
{\small
\begin{tabular}{c|p{2cm}|p{2cm}|p{2cm}}
Sol. & $X_1$ & $X_2$ & $X_3$ \\
\hline
$s_1$ 	& \{C, F, G, H\} 		& \{E\}       		& \{A, D, F\} \\
$s_2$ 	& \{C, F, G, H\} 		& \{A, D, F\}   	& \{E\}     \\
$s_3$ 	& \{A, D, F\}   		& \{C, F, G, H\} 	& \{E\}     \\
$s_4$ 	& \{A, D, F\}   		& \{E\}   		& \{C, F, G, H\}    \\
$s_5$ 	& \{E\} 			& \{C, F, G, H\} 	& \{A, D, F\} \\
$s_6$ 	& \{E\}			& \{A, D, F\}   	& \{C, F, G, H\} \\
$s_7$ 	& \{A, F\} 			& \{C, H\} 		& \{E, G\} 	\\
$\vdots$	& \vdots			& \vdots		& 	\vdots 	\\
$s_{13}$	& \{C, E, H\}		& \{E, G\} 		& 	\{F\} 	\\
$\vdots$	& \vdots		& \vdots		& 	\vdots 	\\
$s_{19}$	& \{A, F\}			& \{C, E, H\}	& 	\{G\}		\\
$\vdots$ 	& \vdots			& \vdots		& 	\vdots 	\\
$s_{25}$	& \{A\}	 		& \{B, E, G\}	& 	\{C\}		\\
$\vdots$	& \vdots		& \vdots	& \vdots \\
$s_{30}$	& \{C\}	 		& \{B, E, G\}	& \{A\} \\
\hline
\end{tabular} 
}
\vspace{0.15cm}
\caption{Set of all solutions (including symmetrical ones).}\label{sols30}
\end{center}
\end{table}

\subsection{Refining queries}

By only refining queries addressing a clustering, the data analyst can easily produce clusterings satisfying different properties. 
In this section, we illustrate this approach by successive refinements.
Starting from initial query $Q_0$, symmetrical solutions are first removed leading to query $Q_1$.
Then, clusterings with non-frequent patterns and clusterings with small size patterns are removed (leading to queries $Q_2$ and $Q_3$).
More generally, this process greatly facilitates the building of global patterns and the discovery of knowledge.

\subsubsection{i) Removing symmetrical solutions.}

Two solutions $s_i$ and $s_j$ are said to be symmetrical iff there exists a permutation $\sigma$, such that $s_j = \sigma (s_i)$.
A clustering problem owns intrinsically a lot of symmetrical solutions:
let $s = (p_1, p_2, ..., p_k)$ be a solution containing $k$ patterns $p_i$.
Any permutation $\sigma$ of these $k$ patterns $\sigma(s) = (p_{\sigma(1)}, p_{\sigma(2)}, ..., p_{\sigma(k)})$ is also a solution. 
So, for any solution, there exist $(k! - 1)$ symmetrical solutions.	
For example, solutions from $s_1$ to $s_6$ are symmetrical (See Table \ref{sols30}) and constitute the same clustering. 

Constraint \texttt{canonical}$([X_1,..., X_k])$ is used to avoid symmetrical solutions.
This constraint states that, for all $i$ s.t. $1 \leq i < k$, pattern $X_i$ is less than pattern $X_{i+1}$ with respect to the lexicographic order.

From query $Q_0$, we obtain query $Q_1$~:
$$
\left\{
\begin{array}{l}
\land_{1 \leq i \leq k}~\texttt{closed}(X_i)~\land\\
\texttt{coverTransaction}([X_1,...,X_k])~\land\\
\land_{1 \leq i < j \leq k}~\texttt{overlapTransactions}(X_i, X_j) = 0 \, \wedge \\
\texttt{canonical}([X_1,...,X_k])
\end{array}
\right .
$$

\noindent Following our running example, query $Q_1$ leads to only 5 solutions since $5$$\times$$3!$$=$$30$ (See Table \ref{sols5}).

The constraint \texttt{canonical}$([X_1,..., X_k])$ plays an important role.
First, as the number of solutions ($k!$) grows very rapidly with the number $k$ of clusters, it quickly becomes very large.
So, it is essential and indispensable to break the symmetries to avoid having a huge number of redundant solutions.
Moreover, this constraint will perform an efficient filtering by drastically reducing the size of the search space.

\begin{table}[t]
\begin{center}
{\small
\begin{tabular}{c|p{2cm}|p{2cm}|p{2cm}}
Sol. & $X_1$ & $X_2$ & $X_3$ \\
\hline
$s_1$ 	& 	\{C, F, G, H\}	& 	\{E\}		& 	 \{A, D, F\} 	\\
$s_7$ 	& 	\{A, F\} 		& 	\{C, H\} 	& 	\{E, G\} 	\\
$s_{13}$ 	& 	\{C, E, H\}		& 	\{E, G\} 	& 	\{F\} 		\\
$s_{19}$ 	& 	\{A, F\}		& 	\{C, E, H\}	& 	\{G\}		\\
$s_{25}$ 	& 	\{A\}	 		& 	\{B, E, G\}	& 	\{C\}		\\
\hline 
\end{tabular} 
}
\vspace{0.15cm}
\caption{Set of different clusterings.}\label{sols5}
\end{center}
\end{table}

\subsubsection{ii) Removing solutions with non-frequent patterns.} 

A clustering containing at least one pattern having a low frequency is not considered to be relevant. 
To remove such solutions, we only need to add new constraints to the current query $Q_1$.
Such a constraint requires that each cluster must have a frequency greater than a threshold (here 10\% of $m=11$). 
$$
\forall \, 1 \le i \le k,~\texttt{freq}(X_i) \ge 2
$$

From query $Q_1$, we obtain query $Q_2$~:
$$
\left\{
\begin{array}{l}
\land_{1 \leq i \leq k}~\texttt{closed}(X_i)~\land\\
\texttt{coverTransaction}([X_1,...,X_k])~\land\\
\land_{1 \leq i < j \leq k}~\texttt{overlapTransactions}(X_i, X_j) = 0 \, \wedge \\
\texttt{canonical}([X_1,...,X_k]) \, \wedge \\
\land_{1 \leq i \leq k}~\texttt{freq}(X_i) \ge 2
\end{array}
\right .
$$

Pattern $\{C, F, G, H\}$ of solution $s_1$ (see Table \ref{sols30}) has a frequency of~$1$ which is less than the threshold.
So for $Q_2$, solution $s_1$ is not valid.
For query $Q_2$, there remain 4 solutions: $s_7$, $s_{13}$, $s_{19}$, and $s_{25}$ (See Table \ref{sols5}).

\subsubsection{iii) Removing solutions with small size patterns.} 

A clustering containing at least one pattern of size $1$ is not considered to be relevant\footnote{Usally, clusterings using these unitary clusters reflect  the discretisation of some attributes.}. 
To remove such clusterings, we only need to add new constraints to the current query $Q_2$.
Such a constraint requires that each cluster must have a size greater than $1$.
This can be acheived by stating, for each cluster, a constraint to restrict its size.

$$
\forall \, 1 \le i \le k,~\texttt{size}(X_i) \ge 2
$$

From query $Q_2$, we obtain query $Q_3$~:
$$
\left\{
\begin{array}{l}
\land_{1 \leq i \leq k}~\texttt{closed}(X_i)~\land\\
\texttt{coverTransaction}([X_1,...,X_k])~\land\\
\land_{1 \leq i < j \leq k}~\texttt{overlapTransactions}(X_i, X_j) = 0 \, \wedge \\
\texttt{canonical}([X_1,...,X_k]) \, \wedge \\
\land_{1 \leq i \leq k}~\texttt{freq}(X_i) \ge 2 \, \wedge \\
\land_{1 \leq i \leq k}~\texttt{size}(X_i) \ge 2
\end{array}
\right .
$$

Query $Q_3$ has only $1$ solution: $s_7$ (see Table \ref{sols5}).
For this solution, we have $X_1 =\{A, F\}$, $X_2=\{C, H\}$ and $X_3= \{E, G\}$.

\subsection{Solving other Clustering Problems}
\label{other-clustering}

In the same way, it is easy to express other clustering problems such as co-clustering, soft clustering and soft co-clustering. 

\subsubsection{i) The \textbf{soft clustering} problem} 
is a relaxed version of the clustering problem where small overlaps (less than $\delta_T$) on transactions are authorized.
This problem is modelised by query $Q_4$ (soft version of $Q_0$):
$$
\left\{
\begin{array}{l}
\land_{1 \leq i \leq k}~\texttt{closed}(X_i)~\land\\
\texttt{coverTransaction}([X_1,...,X_k])~\land\\
\land_{1 \leq i < j \leq k}~\texttt{overlapTransactions}(X_i, X_j) \le \delta_T \, \wedge \\
\texttt{canonical}([X_1,...,X_k])
\end{array}
\right .
$$

Consider query $Q_4$ with $k$$=$$3$ and a maximal overlap for transactions $\delta_T$$=$$1$. 
There are $13$ solutions (see Table \ref{sc_res}).
If symmetries are not broken using the constraint \texttt{canonical}($[X_1,...,X_k]$), then there are $78$ ($3!$$\times$$13$) solutions.

For solution $s'_1$, patterns $X_1$ and $X_3$ cover transaction $t_{11}$ (see Table \ref{tab}).
Moreover, patterns $X_2$ and $X_3$ cover transaction $t_2$ (see Table \ref{tab}).
After having removed solutions with non-frequent patterns, there remain $8$ solutions: 
from $s'_6$
to $s'_{13}$. 
After having removed solutions with small size patterns, it remains only $1$ solution: 
$s'_9$ (which is the solution $s_7$ of the initial clustering problem, see Section \ref{pbsol}).

\begin{table}
\centering
\begin{tabular}{c|l|l|l}
Sol. & $X_1$ & $X_2$ & $X_3$ \\
\hline
$s'_1$  & \{C, F, G, H\} 	& \{E\} 	& \{F\} \\
$s'_2$  & \{A, D, F\} 	& \{C, F, G, H\}	& \{E\} \\
$s'_3$  & \{A, F\} 	& \{C, F, G, H\}	& \{E\} \\
$s'_4$  & \{A, E, F\} 	& \{E\}		& \{F\} \\
$s'_5$  & \{A, D, F\} 	& \{E\}		& \{F\} \\
$s'_6$  & \{A\} 		& \{B, E, G\}	& \{C\} \\
$s'_7$  & \{A, F\} 	& \{C, E, H\}	& \{G\} \\
$s'_8$  & \{A, F\} 	& \{C, H\}	& \{G\} \\
$s'_9$  & \{A, F\} 	& \{C, H\}	& \{E, G\} \\
$s'_{10}$ & \{C, E, H\} 	& \{E, G\}	& \{F\} \\
$s'_{11}$ & \{C, H\} 	& \{E, G\}	& \{F\} \\
$s'_{12}$ & \{C, H\} 	& \{F\}		& \{G\} \\
$s'_{13}$ & \{C, E, H\} 	& \{F\}		& \{G\} \\
\hline
\end{tabular}
\vspace{0.15cm}
\caption{Set of different clusterings for query $Q_4$ (soft clustering). \label{sc_res}}
\end{table}

\subsubsection{ii) The \textbf{co-clustering} problem}
consists in finding $k$ clusters covering both the set of transactions and the set of items, without any overlap on transactions or on items.
This problem is modelised by query $Q_5$:
$$
\left\{
\begin{array}{l}
\land_{1 \leq i \leq k}~\texttt{closed}(X_i)~\land\\
\texttt{coverTransaction}([X_1,...,X_k])~\land\\
\land_{1 \leq i < j \leq k}~\texttt{overlapTransactions}(X_i, X_j) = 0 \, \wedge \\
\texttt{coverItems}([X_1,...,X_k])~\land \\
\land_{1 \leq i < j \leq k}~\texttt{overlapItems}(X_i, X_j) = 0 \, \wedge \\
\texttt{canonical}([X_1,...,X_k])
\end{array}
\right .
$$

\subsubsection{iii) The \textbf{soft co-clustering} problem} is a relaxed version of the co-clustering problem,
allowing small overlaps on transactions (less than $\delta_T$) and on items (less than $\delta_I$).
This problem is modelised by query $Q_6$ (soft version of $Q_4$ and $Q_5$):
$$
\left\{
\begin{array}{l}
\land_{1 \leq i \leq k}~\texttt{closed}(X_i)~\land\\
\texttt{coverTransaction}([X_1,...,X_k])~\land\\
\land_{1 \leq i < j \leq k}~\texttt{overlapTransactions}(X_i, X_j) \le \delta_T \, \wedge \\
\texttt{coverItems}([X_1,...,X_k])~\land \\
\land_{1 \leq i < j \leq k}~\texttt{overlapItems}(X_i, X_j) \le \delta_I \, \wedge \\
\texttt{canonical}([X_1,...,X_k])
\end{array}
\right .
$$

\section{Conclusions and Future Works}
We have proposed a constraint-based language allowing to easily express
different mining tasks in a declarative way. Thanks to the declarative
process, extending or changing the specification to refine the results and get more relevant patterns or address new global patterns is very simple. Moreover, all constraints can be combined together and new constraints can be added.

The effectiveness and the flexibility of our approach is shown on several examples coming from clustering based on associations: thanks to query refinements, the data analyst is able to produce clusterings satisfying different constraints, thus generating more meaningful clusters and avoiding outlier ones.

As future work, we intend to enrich our constraint-based language with further constraints to capture and model a wide range of data mining tasks. 
The scalability of the approach to larger values of $k$ and larger datasets can also be investigated. 
Another promising direction is to integrate optimisation criteria in our framework.

%
%

\bibliographystyle{plain}

\end{document}